\documentclass[twoside,twocolumn]{article}

\usepackage{blindtext} 
\usepackage{amsmath}
\usepackage{amsfonts}
\usepackage{amssymb}
\usepackage{dblfloatfix}    
\usepackage{graphicx}
\graphicspath{ {} }
\usepackage[sc]{mathpazo} 
\usepackage[T1]{fontenc} 
\usepackage{microtype} 

\usepackage[english]{babel} 

\usepackage[hmarginratio=1:1,top=32mm,columnsep=20pt]{geometry} 
\usepackage[hang, small,labelfont=bf,up,textfont=it,up]{caption} 
\usepackage{booktabs} 

\usepackage{lettrine} 

\usepackage{enumitem} 
\setlist[itemize]{noitemsep} 

\usepackage{abstract} 

\usepackage{titlesec} 
\renewcommand\thesection{\Roman{section}} 
\renewcommand\thesubsection{\roman{subsection}} 
\titleformat{\section}[block]{\large\scshape\centering}{\thesection.}{1em}{} 
\titleformat{\subsection}[block]{\large}{\thesubsection.}{1em}{} 

\usepackage{fancyhdr} 
\usepackage{titling} 

\usepackage{hyperref} 

\pretitle{\begin{center}\Huge\bfseries} 
\posttitle{\end{center}} 
\title{Word Embeddings and Their Use In Sentence Classification Tasks} 
\author{%
\textsc{Amit Mandelbaum \hspace{25pt} Adi Shalev}
\\[1ex] 
\normalsize Hebrew University of Jerusalm  \\ 
\normalsize amit.mandelbaum@mail.huji.ac.il \hspace{25pt} bitan.adi@gmail.com 
}
\date{\today} 
\renewcommand{\maketitlehookd}{%
\begin{abstract}
\noindent This paper has two parts. In the first part we discuss word embeddings. We discuss the need for them, some of the methods to create them, and some of their interesting properties. We also compare them to image embeddings and see how word embedding and image embedding can be combined to perform different tasks. In the second part we implement a convolutional neural network trained on top of pre-trained word vectors. The network is used for several sentence-level classification tasks, and achieves state-of-art (or comparable) results, demonstrating the great power of pre-trainted word embeddings over random ones.
\end{abstract}
}

\begin{document}

\maketitle


\section{Introduction}

\lettrine[nindent=0em,lines=3]{T}here are some definitions for what \textit{Word Embeddings} are, but in the most general notion, word embeddings are the numerical representation of words, usually in a shape of a vector in $\Re^d$. Being more specific, word embeddings are unsupervisedly learned word representation  vectors whose relative similarities correlate with semantic similarity. In computational linguistics they are often referred as \textit{distributional semantic model} or \textit{distributed representations}. 
\par The theoretical foundations of word embeddings can be traced back to the early 1950's and in particular in the works of Zellig Harris, John Firth, and Ludwig Wittgenstein. The earliest attempts at using feature representations to quantify (semantic) similarity used hand-crafted features. A good example is the work on semantic differentials \cite{Osgood:semantic}.
The early 1990's saw the rise of automatically generated contextual features, and the rise of Deep Learning methods for Natural Language Processing (NLP) in the early 2010's helped to increase their popularity, to the point that, these days, word embeddings are the most popular research area in NLP \footnote{In 2015 the dominating subject at EMNLP ("Empirical Methods in NLP") conference was word embeddings, source: http://sebastianruder.com/word-embeddings-1/}.
\par This work will be divided into two parts. In the first part we will discuss the need for word embeddings, some of the methods to create them, and some interesting features of those embeddings. We also compare them to image embeddings (usually referred as image features) and see how word embedding and image embedding can be combined to perform different tasks.
\par In the second part of this paper we will present our implementation of Convolutional Neural Networks for Sentence Classification \cite{Kim:convsent}. This work which became very popular is a very good demonstration of the power of pre-trained word embeddings. Using a relatively simple model, the authors were able to achieve state-of-art (or comparable) results, for several sentence-level classification tasks. In this part we will present the model, discuss the results and compare them to those of the original article. We will also extend and test the model on some datasets that were not used in the original article. Finally, we will propose some extensions for the model which might be a good proposition for future work.

\section{Word Embeddings}
\subsection{Motivation}
It is obvious that every mathematical system or algorithm needs some sort of numeric input to work with. However, while images and audio naturally come in the form of rich, high-dimensional vectors (i.e. pixel intensity for images and power spectral density coefficients for audio data), words are treated as discrete atomic symbols. 
\par The naive way of converting words to vectors might assign each word a one-hot vector in $\Re^{|V|}$ where |V| being vocabulary size. This vector will be all zeros except one unique index for each word.
Representing words in this way leads to substantial data sparsity and usually means that we may need more data in order to successfully train statistical models. 
\begin{figure}[!h] 
	\centering
	\includegraphics[width=0.45\textwidth]{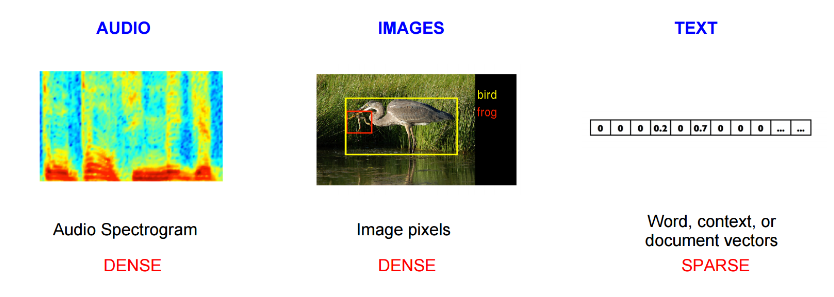}
	\caption{Density of different data sources.}
\end{figure}
\par What mentioned above raise the need for continuous, vector space representations of words that contain data that can be leveraged by models. To be more specific we want semantically similar words to be mapped to nearby points, thus making the representation carry useful information about the word actual meaning.

\subsection{Word Embeddings Methods}
Word embeddings models can be divided into main categories:
\begin{itemize}
	\item Count-based methods
	\item Predictive methods
\end{itemize}
Models in both categories share, in at least some way, the assumption that words that appear in the same contexts share semantic meaning.

\par One of the most influential early works in count-based methods is the LSI/LSA (Latent Semantic Indexing/Analysis) \cite{Darweesher:LSI} method. This method is based on the \textit{Firth's hypothesis} from 1957 \cite{Firth} that the meaning of a word is defined "by the company it keeps". This hypothesis leads to a very simple albeit a very high-dimensional word embedding. Formally, each word can be represented as a vector in $\Re^N$ where N is the unique number of words in a given dictionary (in practice N=100,000). Then, by taking a very large corpus (e.g. Wikipedia), let $Count_5(w_1,w_2)$  be the number of times $w_1$ and $w_2$ occur within a distance 5 of each other in the corpus. Then the word embedding for a word w is a vector of dimension N, with one coordinate for each dictionary word. The coordinate corresponding to word $w_2$ is $Count_5(w,w_2)$. 
\par The problem with the resulting embedding is that it uses extremely high-dimensional vectors. In the LSA article, is was empirically discovered that these embeddings can be reduced to vectors $R^{300}$ by doing a rank-300 SVD on the NxN original embeddings matrix. 

This method was later refined with reweighting heuristics, such as taking the logarithm, or Pointwise Mutual Information (PMI) \cite{Kenneth:PMI} on the count, which is a very popular method.
\par The second family of methods, sometimes also referred as \textit{neural probabilistic language models}, had theoretical and some practical appearance as early as 1986 \cite{Hinton:dist}, but first to show the utility of pre-trained word embeddings were arguably Collobert and Weston in 2008 \cite{Collobert:Deep2008}. Unlike count-based models, predictive models try to predict a word from its neighbors in terms of learned small, dense embedding vectors.
\par Two of the most popular methods which appeared recently are the Glove (Global Vectors for Word Representation) method \cite{Glove}, which is an unsupervised learning method, although not predictive in the common sense, and Word2Vec, a family of energy based predictive models, presented by \cite{Mikolov:word2vec}. As Word2Vec is the embedding method used in our work it shall be briefly discussed here.

\subsection{Word2Vec}
Word2vec is a particularly computationally-efficient predictive model for learning word embeddings from raw text. It comes in two flavors, the Continuous Bag-of-Words model (CBOW) and the Skip-Gram model. Algorithmically, these models are similar, except that CBOW predicts target words (e.g. 'mat') from source context words ('the cat sits on the'), while the skip-gram does the inverse and predicts source context-words from the target words.
\begin{figure}[!h] 
	\centering
	\includegraphics[width=0.45\textwidth]{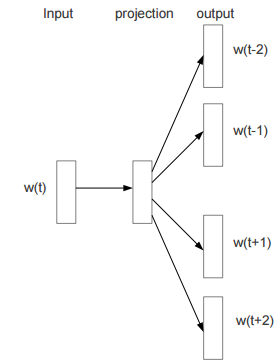}
	\caption{The Skip-gram model architecture}
\end{figure}
\begin{figure*}
	\centering
	\includegraphics[width=1\textwidth]{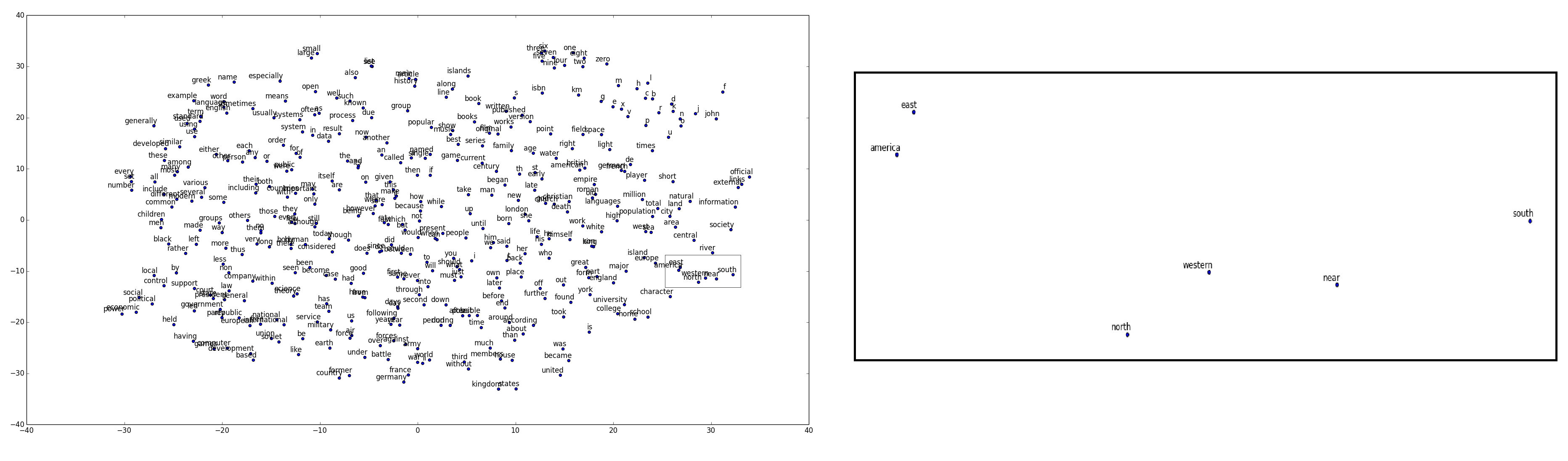}
	\caption{Left: Word2Vec t-SNE \cite{T-sne} visualization of our implementation, using text8 dataset and a window size of 5. Only 400 words are visualized. Right: Zooming in of the rectangle in the left figure.}
\end{figure*}
In the skip-gram model (see figure 2) a neural network is trained over a large corpus in where the training objective is to learn word vector representations that are good at predicting the nearby words.
The method is also using a simplified version of NCE \cite{NCE} called \textit{Negative sampling} where the objective function is defined as follows:
\begin{equation}
	log\sigma(v_{w_O}^{\prime T}v_{w_I})+\sum_{i=1}^{k}\mathbb{E}_{w_i\sim P_n}(w)[\sigma(-v_{w_i}^{\prime T}v_{w_I})]
\end{equation}
where $v'_w$ and $v_w$ are the "input" and "output" vector representations of w, $\sigma$ is the \textit{sigmoid} function but can also be seen as the network parameters function, and $P_n$ is some noise probability used to sample random words.
In the article they recommend k to be between 5 to 20, while the context of predicted words should be 5 or 10.
This above objective is later put in the Skip-Gram objective (equtaion 2) to produce optimal word embeddings.
\begin{equation}
\frac{1}{T}\sum_{t=1}^{T}\sum_{-c\leq j\leq c,j\neq 0}log p(w_{t+j}|w_t)
\end{equation}
This objective enables the model to differentiate data from noise by means of logistic
regression, thus learning high-quality vector representations.
\par The CBOW does exactly the same but the direction is inverted. In other words the CBOW trains a binary logistic classifier where, given a window of context words, gives a higher probability to "correct" if the next word is correct and a higher probability to "incorrect" if the next word is a random sampled one. 
Notice that CBOW smoothes over a lot of the distributional information (by treating an entire context as one observation). For the most part, this turns out to be a useful thing for smaller datasets. However, skip-gram treats each context-target pair as a new observation, and this tends to do better when we have larger datasets.
\par Finally the vector we used in our work had a dimension of 300. The Network was trained on the Google News dataset which contains 30 billion training words, with negative sampling as mentioned above. These embeddings can be found online\footnote{code.google.com/p/word2vec}.
\par A lot of follow-up work was done on the Word2Vec method. One interesting work was done by \cite{Levy:word2Vec_explained} where experiments and theory were used to suggest that these newer methods are related to the older PMI based models, but with new hyperparameters and/or term reweightings.
In this project appendix you can find a simplified version of Word2Vec we implemented in TensorFlow architecture using the text8 dataset\footnote{http://mattmahoney.net/dc/textdata} and the Skip-Gram model. See figure 3 for visualized results.
\subsection{Word Embeddings Properties}
\textbf{Similarity:} The simplest property of embeddings obtained by all the methods described above is that similar words tend to have similar vectors. More formally, the similarity between two words (as rated by humans on a [-1,1] scale) correlates with the cosine similarity between those words' vectors.
\begin{figure}[!h] 
	\centering
	\includegraphics[width=0.5\textwidth]{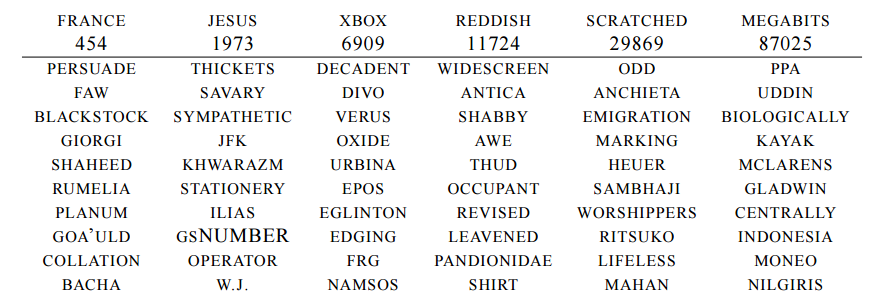}
	\caption{What words have embeddings closest to a given word? From \cite{collobert:scratch}}
\end{figure}
The fact that words embedding are related to their context-words stand behind the similarity property as naturally, similar words tend to appear in similar context. This, however creates the problem that antonyms (e.g. cold and hot etc.) also appear with the same context while they are, by definition, have opposite meaning. In \cite{Mikolov:word2vec} the score of the (accept,reject) pair is 0.73,
and the score of (long,short) is 0.71. 
\par The problem of antonyms was tackled directly by \cite{Schwartzh:symmetric}. In this article, the authors introduce a symmetric pattern based approach to word representation which is particularly suitable for capturing word similarity. Symmetric patterns are a special type of patterns that contain exactly two wildcards and that tend
to be instantiated by wildcard pairs such that each member of the pair can take the X or the Y position. For example, the symmetry of the pattern "X or Y" is exemplified by the semantically plausible expressions "cats or dogs" and "dogs or cats".
Specifically it was found that two patterns are particularly indicative of antonymy - "from X to Y" and "either X or Y".
\par Using their model the authors were able to achieve a  $\rho$ score of 0.56 on the simlex999 dataset \cite{simlex999}, improving state-of-the-art word2vec skip-gram model results by as much as 5.5-16.7\%. Furthermore, the authors demonstrated the adaptability
of their model to antonym judgment specifications.
\par \textbf{Linear analogy relationships:} A more interesting property of recent embeddings \cite{Mikolov:word2vec} is that they can solve analogy relationships via linear algebra. This is despite the fact that those embeddings are being produced via nonlinear methods.
For example, $v_{queen}$ is the most similar answer to the $v_{king}-v_{men}+v_{women}$ equation. It turns out, though, that much more sophisticated relationships are also encoded in this way as we can see in figure 5 below.
\begin{figure}[!h] 
	\centering
	\includegraphics[width=0.5\textwidth]{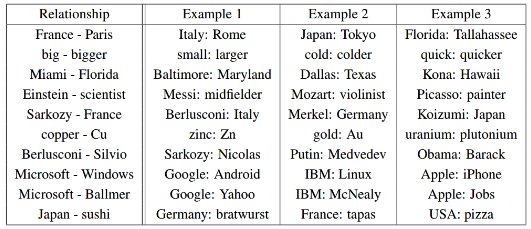}
	\caption{Relationship pairs in a word embedding. From \cite{Mikolov:word2vec}}
\end{figure}
\par An interesting theoretical work on non-linear embeddings (especially PMI) was done by \cite{Rand-walk}. In their article they suggest that the creation of a textual corpus is driven by the random walk of a discourse vector $c_t \in \Re^d$, which is a unit vector whose direction in space represents what is being talked about.
Each word has a (time-invariant) latent vector  $v_w \in \Re^d$ that captures its correlations with the discourse vector. Using a word production model they predict that words occurring at successive time steps will also tend to have vectors that are close together, thus explaining why similar words have similar vectors. 
\par Using the above model the authors introduce the "RELATIONS = DIRECTIONS" notion for linear analogies. The authors claim that for each relation R, some direction $\mu_R$ can be found which satisfies some equation. This leads to the finding that given enough examples of a relationship R, it is possible to compute $\mu_R$ using SVD and then given a pair of words with a realtion R and a word c, find the best analogy with word d by finding the pair c and d such that $v_c-v_d$ has highest possible projection over $\mu_R$. In this way, thay also explain that low dimension of the vectors has a "purifying" effect that reduces the effect of the overfitting coming from the PMI approximation, thus achieving much better results than higher dimensional vectors.

\subsection{Word Embeddings Extensions}
In this last subsection we will review two interesting works that extend the word embedding concept to phrases and sentences using different approaches.
\par In \cite{Mitchell:Semantic} the authors address the problem that vector-based models are typically directed at representing words in isolation and methods for constructing representations for phrases or sentences have received little attention in the literature. The authors suggests the use of two composition operations, multiplication and addition (and their combination). This way the authors are able to combine word embeddings into phrase or sentences embeddings while taking into account important properties like word order and semantic relationship between words (i.e. semantic composition types).
\par In MIL (Multi Instance Transfer Learning)  \cite{MIL} the authors propose a neural network model which learns embedding at increasing level of hierarchy, starting from word embeddings, going to sentences and ending with entire document embeddings. The authors then use transfer  learning by pulling the sentence or word embedding that were trained as part of the document embeddings and use them for sentence or word review classification or similarity tasks (See figure 6 below).
\begin{figure}[!h] 
	\centering
	\includegraphics[width=0.45\textwidth]{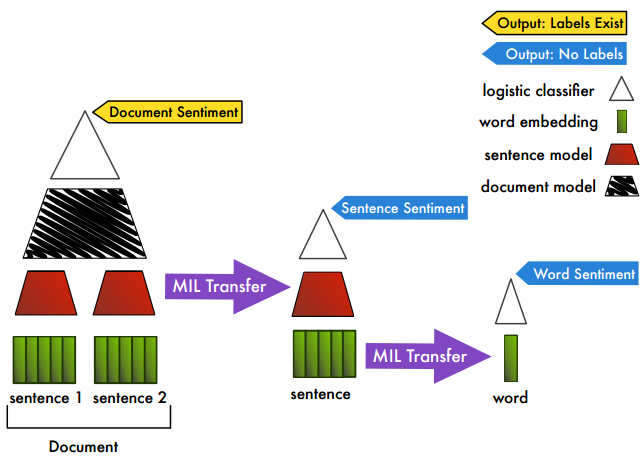}
	\caption{Deep multi-instance transfer learning approach for review data, taken from \cite{MIL}}
\end{figure}

\section{Word Embeddings vs. Image Embeddings}
\subsection{Image Embeddings}
Image embeddings, or image features, were wildly used for most image processing and classification tasks until the early 2010's. The features ranged from simple histograms or edge maps to the more sophisticated and very popular SIFT \cite{SIFT} and HOG \cite{HOG}. However, recent years have seen the rise of Deep Learning for image classification, especially since 2012 when the AlexNet \cite{AlexNet} article was published. As those Convolutional Neural Networks (CNN) operated directly on the images, it was suggested that these networks learn the best image features for the specific task that they are trained for, thus obviating the need for specific hand-crafted features.
\begin{figure}[!h] 
	\centering
	\includegraphics[width=0.5\textwidth]{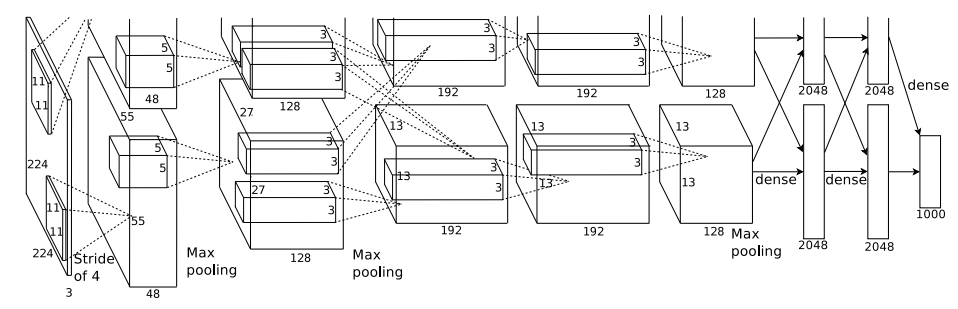}
	\caption{The CNN architecture of AlexNet}
\end{figure}

\begin{figure*}[!b]
	\centering
	\includegraphics[width=1\textwidth]{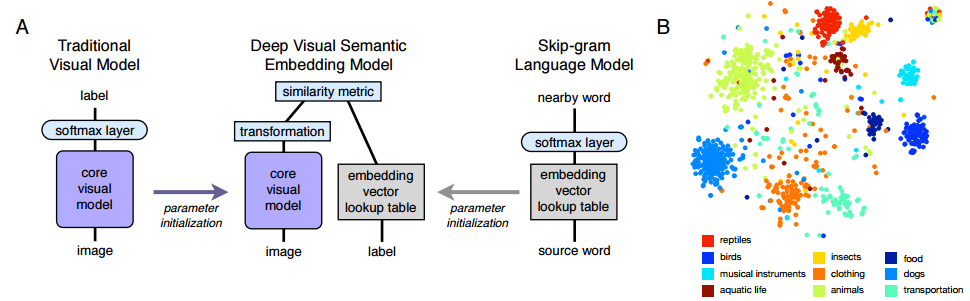}
	\caption{: (a) Left: a visual object categorization network with a softmax output layer; Right: a skip-gram language model; Center: the joint model, which is initialized with parameters pre-trained at the lower layers of the other two models. (b) t-SNE visualization [19] of a subset of the ILSVRC 2012 1K label embeddings learned using skip-gram. Taken from \cite{DeVise}}
\end{figure*}

\par In recent years though, an extensive research was done on the nature and usage of the kernels and features learned by CNN's. Extensive study of CNN feature layers was done in \cite{Visualizing convnets} where they empirically confirmed that each convolutional layer of the CNN learns a set of filters. Their experiments also confirm that filters complexity and expressive power is rising from layer to layer (i.e. as the network goes deeper) starting from simple edge detectors to complex objects detectors like eyes, flowers, faces and more.
The authors also suggest using the pre-trained one before last layer as a feature map, or \textit{Image Embeddings} input for simpler SVM classifiers. 
\par Another popular work was done a bit earlier in \cite{Decaf} where they also used a pre-trained CNN features as a base for visual recognition tasks.
This work was followed by several works with one of them being considered the philosophical father of the algorithm we implement later. In \cite{Cnn features off the shelf} the authors used the one before last layer of a network similar to AlexNet that was pre-trained on ImageNet \cite{ImageNet} as image embeddings. The authors were able to acheive state-of-art results on several recognition tasks, using simple classifiers like SVM. The result was surprising due to the fact that the CNN model was originally optimized for the task of object classification in ILSVRC 2013 dataset. Nevertheless, it showed itself to be a strong competitor to the more sophisticated and highly tuned state-of-the-art methods. 
\par These works and others suggested that given a large enough database of images, a CNN can learn an image embedding which captures the "essence" of the picture and can be used later as an input to different tasks, similar to what is done with word embeddings. 
\subsection{Similarities and Differences}
As we saw earlier Word embedding and Image embeddings are similar in the sense that while they are being learned as part of a specific task, they can be successfully used later for a variety of other tasks. Also, in both cases, similar images or words will usually have similar embeddings.
However Word embeddings and image embeddings differ in some aspects. 
\par The first difference is that while word embeddings depends mostly on the words surrounding the given word, image embeddings usually rely on the specific image itself. This might explain the fact that linear analogies does not appear naturally in images. An interesting work was done in \cite{Visual analogy making} where a neural network is trained to make visual analogies and learn to make them based on appearance, rotation, 3D pose, and various object attributes.
\par Another difference is that while word embeddings are usually low-ranked, image embeddings might have same or even higher dimension then the original image. Those embeddings are still useful as they contain a lot of information that is extracted from the image and can be used easily. 
\par Lastly, we notice that word embeddings are trained on a specific corpus where the final embedding results come as the form of word-vectors. This limits the embedding to be valid only for words that were found in the original corpus while other words will need to be initialized as random vectors (as also done in our work). In images on the other hand, the embeddings come as a pre-trained model where features or embeddings can be pulled for any sort of image by feeding the model with the image, making image embeddings models a bit more robust (although they might subjected to other constraints like size and image type).
\begin{figure*}
	\centering
	\includegraphics[width=0.9\textwidth]{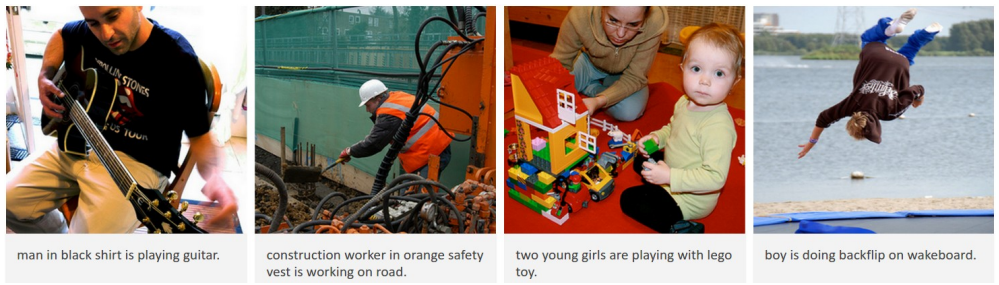}
	\caption{Image captiones generated with Deep Visual-Semantic model Taken from \cite{Deep visual semantic=}}
\end{figure*}
\subsection{Joint Word-Image Embeddings}
To conclude this part we will review some of the recent work done in the exciting area of joint word-image embeddings.
The first immediate usage of joint word-image embeddings is image annotations or image labeling. An early notable work was done by \cite{Wetson:image annotation } where a representation of images and representation of annotations where both mapped to a joint feature space by learning a mapping which optimizes top-of-the-list ranking measures for images and annotations. This method, however, learns linear mappings from image features to the embedding space, and the available labels were only those provided in the image training set. It could thus not generalize to new classes.
\par In 2014 DeVise (A Deep Visual-Semantic Embedding Model) model was shown by \cite{DeVise}. This work which continued earlier work \cite{Socher: image annotation}, combined image embedding and word embedding trained \textbf{separately} into joint similarity metric (see figure 8).
This enabled them to give performance comparable to a state-of-the-art softmax based model on a flat object classification metric, while simultaneously making more semantically reasonable errors. Their model was also able to make correct predictions across thousands of previously unseen classes by leveraging semantic knowledge elicited only from un-annotated text.
\par Another line of works which combines image and words embeddings is the image captioning area. In this area the embeddings are usually not combined into a joint space but rather used together to create captions for images. In \cite{Deep visual semantic=} an image features pulled from a pre-trained CNN are fed into a Recurrent Neural Network (RNN) which uses word embeddings in order to generate a captioning for the image, based on the image features and previous words (see figure 9).
This sort of combination appears in most image captioning works or video action recognition tasks. 
\par Finally, a slightly more sophisticated method combining RNN's and Fisher Vectors can be found in \cite{LiorWolf} where the authors were able to achieve state-of-art results on both image captioning and video action recognition tasks, using transfer learning on the embeddings learned for the image captioning tasks.  

\section{CNN for Sentence Classification Model}
In this section and the following we are going to represent our implementation of The Convolutional Neural Networks for Sentence Classification model \cite{Kim:convsent} and our results.
This model gained much popularity since it was first introduced in late 2014, mainly because it provides a very strong demonstration for the power of pre-trained word embeddings.
\par The model and results were examined in detail in \cite{CnnConvSentAnalysis} where they test many types of 
configurations for the model, including different sizes and number of filters, different activation units and different word embbeddings. 
\par A partial implementation of the model was done in Theano framework by the authors\footnote{https://github.com/yoonkim/CNN\_sentence} and another simplified version of the model was done in TensorFlow\footnote{https://github.com/dennybritz/cnn-text-classification-tf}. In our work we used small parts of the mentioned codes, however most of the code had to be re-written and expanded in order to perform a true implmentation of the article's model.
\subsection{Model details}
The model architecture, shown in figure 10, is a slight variant of the CNN architecture of \cite{collobert:scratch}. 
\begin{figure*}
	\centering
	\includegraphics[width=0.9\textwidth]{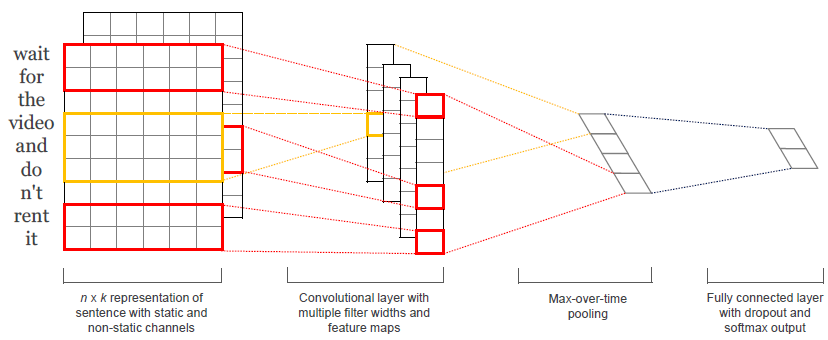}
	\caption{Model architecture with two channels for an example sentence. Taken from \cite{Kim:convsent}}
\end{figure*}
Formally, let $x_i \in \Re^k$ be the k-dimensional word vector corresponding to the \textit{i}-th word in the sentence. Let n be the length (in number of words) of the longest sentence in the dataset, and let $l_h$ be the width of the widest filter in the network. Then, the input to the network is a $k\times (n+l_h-1)$ matrix, which is a concatenation of the word embeddings vectors of each sentences, padded by $l_h-1$ zero vectors in the beginning and some more zero vectors in the end so there are $n+l_h-1$ vectors eventually.
\par The input of the network is convolved with filters of different widths (i.e. number of words in the window) and different sizes (i.e. number of features). For example, a feature $c_i$ is generated from a window of words $x_{i:i+h-1}$ by a filter with width \textit{h} is:
\begin{equation}
c_i = f(wx_{i:i+h-1}+b)
\end{equation}
where \textit{w} are the filter weights, \textit{b} is a bias term and \textit{f} is a non-linear function like ReLU. This process is done for all filters and for all words to create a number of feature maps for each filter.
Next, those features maps are then max-pooled (so we can deal with different sentence sizes) and finally connected to a soft-max classification layer.
\par For regularization we employ dropout \cite{DropOut} on the penultimate layer. This entails randomly (under some probability) setting values in the weight vector to 0. In the original article they also employed constraint on $l_2$ norms of this layer, however \cite{CnnConvSentAnalysis} found that it had negligible contribution to results and therefore was not used here.
\par Training of the network is done by minimizing the corss-entropy loss between predicted labels (soft-max layer) and correct ones. The parameters
to be estimated include the weight vector(s), of the filter(s), the bias term in the activation function, the weight vector of the softmax function and (optionally) the word embeddings. Optimization is performed using
SGD \cite{SGD} and back-propagation, with a small mini-batch size specified later.

\section{Datasets}
We test our model on various benchmarks. Some of them used in the original article while others are extension we do to the original work. You can see the dataset statistics summary in table 1 below.

\begin{itemize}
	\item \textbf{MR:} Movie reviews with one sentence per review.
	Classification involves detecting positive/negative reviews \cite{MR dataset}\footnote{https://www.cs.cornell.edu/people/pabo/movie-review-data/}
	\\
	\item \textbf{SST-1:} Stanford Sentiment Treebank-an
	extension of MR but with train/dev/test splits
	provided and fine-grained labels (very positive,
	positive, neutral, negative, very negative), re-labeled by \cite{Socher-SST-1} \footnote{http://nlp.stanford.edu/sentiment/ Data is actually provided at the phrase-level and hence we train the model on both phrases and sentences but only score on sentences at test time, as in \cite{Socher-SST-1} Thus the training set is an order of magnitude larger than listed in table 1.}
	\\
	\item \textbf{SST-2:} Same as SST-1 but with neutral reviews
	removed and binary labels.
	\\
	\item \textbf{Subj:} Subjectivity dataset where the task is
	to classify a sentence as being subjective or
	objective \cite{Subj}.
	\\
	\item \textbf{TREC:} TREC question dataset-task involves
	classifying a question into 6 question
	types (whether the question is about person,
	location, numeric information, etc.) \cite{TREC}\footnote{http://cogcomp.cs.illinois.edu/Data/QA/QC/}.
	\\
	\item \textbf{Irony: }\cite{Irony} This
	contains 16,006 sentences from reddit labeled as
	ironic (or not). The dataset is imbalanced (relatively
	few sentences are ironic). Thus before training,
	we under-sampled negative instances to make
	classes sizes equal. This dataset was not used in the original article but was tested in \cite{CnnConvSentAnalysis}.
	\\
	\item \textbf{Opi:} Opinions dataset, 	which comprises sentences extracted from user reviews 	on a given topic, e.g. "sound quality of ipod nano". There are 51 such topics and each topic 	contains approximately 100 sentences. The test is to classify which opinion belongs to which topic \cite{Opi}. This dataset was not used in the original article but was tested in \cite{CnnConvSentAnalysis}.
	\\
	\item \textbf{Tweet:} Tweets from 10 different authors. Classification involves classifying which Tweet belongs to which author\footnote{http://www.cs.huji.ac.il/~nogazas/pages/projects.html, Thanks to Noga Zaslavsky}. This dataset was not used in the original article.
	\\
	\item \textbf{Polite:} Sentnces taken Wikipedia editors' logs which have 25 ranges of politeness \cite{Polite}. We narrowed it to 2 binary classes (polite/inpolite).  This dataset was not used in the original article.
	
\end{itemize}

\begin{table}[]
	\centering
	
	\label{my-label}
	\begin{tabular}{|l|l|l|l|l|l|l|}
		\hline
		Data      & c  & l  & N     & |V|   & $|V_{pre}|$ & Test \\ \hline
		MR        & 2  & 20 & 10662 & 18765 & 16488    & CV   \\ 
		SST-1     & 5  & 18 & 11855 & 17836 & 16262    & 2210 \\ 
		SST-2     & 2  & 19 & 9613  & 16185 & 14838    & 1821 \\ 
		Subj      & 2  & 23 & 10000 & 21323 & 17913    & CV   \\ 
		TREC      & 6  & 10 & 5952  & 9592  & 9125     & 500  \\ 
		Irony     & 2  & 75 & 1074  & 6138  & 5718     & CV   \\ 
		Opi       & 51 & 38 & 7086  & 7310  & 6538     & CV   \\ 
		Tweet     & 10 & 39 & 25552 & 33438 & 17023    & 5964 \\ 
		Polite    & 2  & 53 & 4353  & 10135 & 7951     & CV   \\ \hline
	\end{tabular}
	\caption{Summary statistics for the datasets after tokenization.
		c: Number of target classes. l: Average sentence length.
		N: Dataset size. |V|: Vocabulary size. |$V_{pre}$|: Number of
		words present in the set of pre-trained word vectors. Test:
		Test set size (CV means there was no standard train/test split
		and thus 10-fold CV was used).}
\end{table}

\begin{table*}[t]
	\centering
	\label{my-label}
	\begin{tabular}{|l|l|l|l|l|l|l|l|l|l|l|}
		\hline
		& \multicolumn{2}{l|}{MR} & \multicolumn{2}{l|}{SST-1} & \multicolumn{2}{l|}{SST-2} & \multicolumn{2}{l|}{Subj} & \multicolumn{2}{l|}{TREC} \\ \hline
		Model & Orig       & Ours       & Orig         & Ours        & Orig         & Ours        & Orig        & Ours        & Orig        & Ours        \\ \hline
		CNN-rand       & 76.1       & 76.4       & 45.0         & 41          & 82.7         & 80.2        & 89.6        & 91.1        & 91.2        & 97.6        \\ 
		CNN-static     & 81.0       & 80.3       & 45.5         & \textbf{48.1}       & 86.8         & 85.4        & 93.0        & 92.5        & 92.8        & 98.2        \\ 
		CNN-non-static & \textbf{81.5}    & 80.5       & 48.0         & 47.3        & \textbf{87.2}   & 85.8        & \textbf{93.4}        & 93          & 93.6        & \textbf{98.6}       \\ \hline
	\end{tabular}
	\caption{Results on datasets that were tested in \cite{Kim:convsent} (Orig above)}.
\end{table*}

\section{Experimental Setup}
\subsection{Hyperparameters and Training}
In our implementation of the model we experimented with a lof of different configurations. Eventually, since results differences were minor, we decided to use the same architecture and parameters mentioned in the original article for all experiments, with some changes mentioned below.
Below is a list of parameters and specifications that were used for all experiments: 
\begin{itemize}
	\item \textbf{Word embeddings:} We used the pre-trained Word2Vec \cite{Mikolov:word2vec} mentioned earlier. Each word embedding is in $\Re^{300}$. For words that are not found in Word2Vec we randomly initialized them with a uniform distribution in the range of [-0.5,0.5].
	\\
	\item \textbf{Filters:} We used filters with window sizes of [3,4,5] with 100 features each. For activation we used ReLU.
	\\
	\item \textbf{Dropout rate:} 0.5.
	\\
	\item \textbf{Mini-Batch size:} 50.
	\\
	\item \textbf{Optimizer:} While AdaDelte optimizer \cite{Adadelta} was used in the original article. We decided to use the more recent ADAM optimizer \cite{ADAM}, as it seemed to converge much faster (i.e. needed less epochs on training) and in some cases improved the results.
	\\
	\item\textbf{Learning rate:} 0.001. We lower it to 0.0005 after 8 epchs and to 0.00005 after 16 epochs. Notice that the original article didn't mention the learning rate.
	\\
	\item\textbf{Number of epochs:}	This was also not mentioned in the original article but can be found in the authors code\footnote{We note here that in the original article they used early stopping with a dev set. However, the early stopping parameters are not mentioned and experiments demanded a lot of coding which is behind the scope of this project. We do assume that the 25 number used in the code might be close enough to the actual number used in the article.}.
	We used 25 epochs for the \textbf{static version} (see Model Variations below). For \textbf{non static} we used either 4 (MR, SST1, SST2, Subj), 10 (Polite), 16 (Twitter, Opi), or 25 (TREC). For the \textbf{random}  version we used 25 except for Tweet where we used 10, and MR and SST-1 where we used 4.
	\\
	\item \textbf{l2-loss:} We added l2-loss with $\lambda=0.15$ on the weights and biases of the final layer. Altough this was not done in the original artcle, we found it to slightly improve the results. 
\end{itemize}

As mentioned earlier, we decided not to use $l_2$ constrains on the norms due to their negligible contribution. 

\subsection{Model Variations}
We experiment with several variants of the model like in the original article.

\begin{itemize}
	\item \textbf{CNN-rand:} Our baseline model where all words are randomly initialized and then modified during training.
	\\
	\item \textbf{CNN-static:} model with pre-trained
	vectors from word2vec. All words-including the unknown ones that are randomly initialized-are kept static and only the other parameters of the model are learned.
	\\
	\item \textbf{CNN-non-static:} Same as above but the pretrained
	vectors are fine-tuned for each task.
\end{itemize}

The authors also used a multi-channel model where one channel is static and the other is not. However, experiments showed that on most datasets, this did not improve the results. As implementing this would have required a lot more coding, we decided to drop it.

\section{Results and discussion}
In this section we will compare the results we got in our implementation to the ones achieved in the original article. Full results can be found in the original article, and we do note that most of them are state-of-art results, or comparable. For datasets that were not present in the original article we shall compare with other achieved results, whether ours or others'. 
\par In table 2 above we can see a comparison between our results and the ones in the original article \cite{Kim:convsent}. We can see that overall, our results are comparable (and sometimes better) to the ones in the original article. We also see that like in the original article, our baseline model with all randomly initialized words (CNN-rand) does not perform well on its own (on most cases). These results suggest that the pre-trained vectors are good, 'universal' feature extractors and can be utilized across datasets. Finetuning the pre-trained vectors for each task gives still further improvements (CNN-non-static).
\par The differences in some of our results can be related to the different optimizer we used, and the fact that we did not use early stopping. We do note that our results (at least on the non-static version) were achieved with much less training than the original article\footnote{That, if we take the 25 epochs in the code we mentioned earlier as an indication to the nubmer of epochs training used in the original article}. We also note that on the \textbf{TREC} dataset we were able to achieve a new state-of-art results, improving the current one ($95\%$) by $3.6\%$. Both these benefits can be related to the use of ADAM \cite{ADAM} optimizer. 
\par On table 3 we can see our results for datasets that were not used in the original article. We also compare them to other results where applicable. 
\begin{table}[]
	\centering
	\label{my-label}
	\begin{tabular}{|l|l|l|l|l|}
		\hline
		Model      & Opi                    & Irony                  & Tweet                     & Polite                 \\ \hline
		Random     & 64.8                   & 60.2                   & 89.1                      & \textbf{66.2}                   \\ 
		Static     & 65.3                   & 60.5                   & 83.8                      & \textbf{66.2}                   \\ 
		Non-Static & \textbf{66.4}                   & 62.1                   & 89.2                      & 65.7                   \\ \hline
		ConvSent   & 64.9                   & 67.1\footnote{This is AUC, not accuracy}             & \multicolumn{1}{c|}{-}    & \multicolumn{1}{c|}{-} \\ \hline
		SVM+TF-IDF  & \multicolumn{1}{c|}{-} & \multicolumn{1}{c|}{-} & \multicolumn{1}{c|}{\textbf{92.5}} & \multicolumn{1}{c|}{-} \\ \hline
	\end{tabular}
	\caption{Resutls for datasets that were not used in the original article. \textbf{Convesent} is \cite{CnnConvSentAnalysis}.}
\end{table}

On the \textit{Opi} and \textit{Irony} dataset we note that the general line of improved results with pre-trained vectors is maintained. On the \textit{Opi} dataset we were also able to achieve a new state-of-art result. We were also able to achieve comparable results on the Irony dataset. Notice that the other reported result is AUC and not accuracy. 
\par The other two results are interesting. On the \textit{Tweet} dataset we notice that random vectors actually perform a lot better than pre-trained static ones. The reason is that on this dataset, almost half of the vocabulary was not found in the Word2Vec embeddings. This makes sense, as tweets usually contain a lot of marks (for example :-) ) and hashtags which would naturally will not be available in embeddings that were trained on news.
This makes the static version a bad choice as it keep the embeddings random during training. 
\par On this dataset we also applied a simple SVM classifier on the TF-IDF features of each tweet. This simple classifier produced much better results, as TF-IDF features are sensitive to uniqe words in a tweet (like hashtags), that usually indicates which is the author, thus making classification easier.
\par On the \textit{Poilte} dataset we notice that results does not matter on the choice of model. The results themselves are also not very good. This results needs further inspection but they might suggests that this model is not fitted for this task or that politeness is a complicated task for automatic classification.

\section{Conclusions and Future Directions}
In this work we reviewed word embeddings. We saw their origins, discussed the different methods for creating them, and saw some of their interesting properties. We think that word embeddings is a very exciting topic for both research and applications, and expect that a lot research is going to be carried for better understanding of their properties and better creation methods. 
\par In this work we also compared Image features and word embeddings and saw how they can be combined to build learning algorithms that can finally gain a good understanding of pictures and scenes. This area is just in its beginning and we expect a lot of work to be carried towards creating a hybrid system which gains understanding of both vision and language, and that combines those understandings together to the benefit of both fields.
\par Finally, we saw that despite little tuning
of hyperparameters, a simple CNN with one layer of convolution, trained on top of Word2Vec embeddings, performs remarkably well on sentence classification tasks. These results add to the well-established evidence that unsupervised pre-training of word vectors is an important ingredient in deep learning for NLP.
\par To conclude this work we propose here two lines for future work that we think might be interesting to check. First, in the spirit of \cite{MIL}, we notice that in our network, the one before last layer is actually learning \textbf{sentence embeddings}. It might be interesting to train the network on some classification task with a relatively large dataset, and then use the learned sentence embeddings in the same fashion word embeddings are used in our work. For example we can train the network on the MR task and then take the learned sentence embeddings and use them as an embedding input to some document classification task. We can then check if this method achieves improvement  over models that try to classify documents using only pre-trained word embeddings.
\par The second line of research is in the spirit of \cite{Visualizing convnets}. ConvNets visualization helped to gain a lot of insights about image sturctre and how features in increasing level of complexity are combined to create images. It might be interesting to apply those same method of visualization to the filters used in our, or similar works and see if the ConvNet filters learn some interesting semantic properties or compositions that can give insights on the structure of language and how computers (or even humans) percept them.
\par 



\end{document}